\documentclass{article}
\usepackage{spconf,amsmath,graphicx}
\usepackage{booktabs} 
\usepackage{multirow, makecell} 
\usepackage{enumitem} 
\usepackage{xspace} 

\usepackage{hyperref} 
\hypersetup{ colorlinks=true, linkcolor=blue, filecolor=magenta, urlcolor=cyan, }

\newcommand{\dataset}{K-hairstyle\xspace}
\newcommand{\datasetNumber}{500,000\xspace}
\newcommand{\trainNumber}{80,966\xspace}
\newcommand{\validationNumber}{23,580\xspace}
\newcommand{\testNumber}{25,306\xspace}
\newcommand{\classNumber}{31\xspace}
\newcommand{\additionalNumber}{63\xspace}
\usepackage{kotex}

\title{K-Hairstyle: A Large-scale Korean hairstyle dataset \\ for virtual hair editing and hairstyle classification}
\name{\begin{tabular}{c}
     Taewoo Kim$^{1}$\textsuperscript{*} \quad Chaeyeon Chung$^{1}$\textsuperscript{*} \quad Sunghyun Park$^{1}$\textsuperscript{*} \quad Gyojung Gu$^{2}$ \\
     Keonmin Nam$^{2}$ \quad Wonzo Choe$^{3}$ \quad Jaesung Lee$^{3}$ \quad Jaegul Choo$^{1}$
\end{tabular}}
\address{$^{1}$KAIST \qquad $^{2}$Nestyle \qquad $^{3}$Brandi}

\newcommand\blfootnote[1]{%
  \begingroup
  \renewcommand\thefootnote{}\footnote{#1}%
  \addtocounter{footnote}{-1}%
  \endgroup
}

\begin{document}

\maketitle

\blfootnote{\textsuperscript{*} These authors contributed equally.}

\begin{abstract}

The hair and beauty industry is a fast-growing industry. This led to the development of various applications, such as virtual hair dyeing or hairstyle transfer, to satisfy the customer needs.
Although several hairstyle datasets are available for these applications, they often consist of a relatively small number of images with low resolution, thus limiting their performance on high-quality hair editing.
In response, we introduce a novel large-scale Korean hairstyle dataset, \dataset, containing \datasetNumber high-resolution images.
In addition, \dataset includes various hair attributes annotated by Korean expert hairstylists as well as hair segmentation masks.
We validate the effectiveness of our dataset via several applications, such as hair dyeing, hairstyle transfer, and hairstyle classification. \dataset is publicly available at~\href{https://psh01087.github.io/K-Hairstyle/}{https://psh01087.github.io/K-Hairstyle/}.
\end{abstract} 

\section{Introduction}
\label{sec:intro}



Hairstyle has long been recognized as a means of expressing one's personality, including age, social status, and racial identification. Therefore, the demand for new technologies, such as virtual hair dyeing or hairstyle transfer has increased to satisfy the consumers.

Recently, several hairstyle datasets, such as Figaro 1k~\cite{figaro2016svanera} and Hairstyle 30k~\cite{yin2017hairstyle}, have been released for applications of deep learning models in the hair and beauty industry. 
However, existing hair datasets often have the limitations as follows.
First, the image resolution of the existing hair datasets is low compared to the resolution of those photos typically taken by the people.
This discrepancy in image resolution causes the models to yield unsatisfactory results.
Second, the number of images in these datasets is often too small for the model to properly learn diverse hairstyles.

In response, we release a novel large-scale Korean hairstyle dataset, \textit{\dataset}, which contains \datasetNumber high-resolution images, hair segmentation masks, and various attribute labels.
\dataset includes ultra-high resolution (\textit{i.e.,} 4032$\times$3024) images that have never been collected before for hairstyle datasets. 
With a lot larger number of images than any other existing datasets, our dataset covers diverse hairstyles, which can lead to a generation of high-resolution images that reflect the details of various hairstyles. The hairstyle categories of \dataset consist of popular hairstyles in Korea. In addition, our dataset does not only contain detailed labels of each hairstyle (e.g., hair length and color) but also has demographic attribute labels (e,g, age and gender). Furthermore, the annotated hair segmentation masks in \dataset can be utilized for hair dyeing and hairstyle transfer. In order to ensure people's anonymity, the faces are blurred.

We demonstrate the effectiveness of \dataset via the following three applications: 1) hair dyeing, 2) reference-based hairstyle transfer, and 3) hairstyle classification. First, we transfer a hair color of an image to the desired color with hair segmentation masks. Additionally, we transfer a hairstyle of an image by leveraging a reference-based image-to-image translation model. Lastly, we conduct an image classification based on various hairstyle attributes labeled by experts.
Even though the faces in \dataset are blurred, we show that hair dyeing and hairstyle transfer models works well in practice.





\begin{table*}[t!]
    \centering
    \begin{tabular}{l|cccccc}
        \toprule
         Dataset & \# images & \# attributes & face blur & resolution & hair segmentation & multi-view \\
        \midrule
        Figaro 1k~\cite{figaro2016svanera} & 1,050 & 7 & No & 250$\times$250 & Yes & No \\
        Hairstyle 30k~\cite{yin2017hairstyle} & 30,000 & 64 & No & 128$\times$128 & No & No \\
        \midrule
        LFW parts~\cite{kae2013augmenting} & 2,927 & N/A & No & 250$\times$250 & Yes & No \\
        CelebAMask-HQ~\cite{lee2020maskgan} & 30,000 & N/A & No & 1024$\times$1024 & Yes & No \\
        \midrule
        \textbf{K-hairstyle (Ours)}& \textbf{\datasetNumber} & \textbf{\classNumber + \additionalNumber} & Yes & \textbf{4032$\times$3024} & Yes & \textbf{Yes} \\
        \bottomrule
    \end{tabular}
    \vspace{-0.2cm}
    \caption{Comparison between \dataset and the existing hairstyle datasets.}
    \vspace{-0.4cm}
    \label{Table:dataset}
\end{table*}

\vspace{-0.2cm}
\section{Related work}

In this section, we compare \dataset with the existing datasets that can be utilized in hairstyle-related tasks. Table~\ref{Table:dataset} summarizes the key differences.

Figaro 1k~\cite{figaro2016svanera} consists of 1,050 images, equally distributed in 7 hairstyle classes. Since this dataset contains hair segmentation masks, diverse angles, and backgrounds, it can be applied to hair detection and segmentation.
However, due to the limited number of hairstyle classes, it cannot be generalized to those tasks involving various hairstyles. 
Additionally, hairstyle 30k~\cite{yin2017hairstyle} is composed of about 30,000 images with the size of 128$\times$128 including 64 different hairstyles. 
Although hairstyle 30k covers relatively more diverse hairstyles than Figaro 1k, the dataset lacks hair segmentation masks, which are important for high-quality hair dyeing or hairstyle transfer tasks.

LFW parts~\cite{kae2013augmenting}, the subset of LFW dataset~\cite{huang2014lfw}, consists of 2,927 facial images annotated with segmentation masks of hair, skin, and background. 
Similarly, CelebA Mask-HQ~\cite{lee2020maskgan} contains 30,000 facial images selected from CelebA~\cite{liu2014celeba}. It also has 512$\times$512 segmentation masks with 19 facial attributes.
However, since these two datasets do not have hairstyle class labels, they are hardly applicable for tasks such as hairstyle classification and transfer.
Furthermore, the maximum number of images in the four datasets mentioned above is 30,000 and the image resolution is 1024$\times$1024 at most.
Also, none of them includes those images captured at various angles (i.e., multi-view images) for each individual.

Unlike the existing datasets, \dataset provides \datasetNumber images with the maximum resolution of 4032$\times$3024, which consist of multi-view images for each individual.
Moreover, the images in \dataset are annotated by \classNumber types of hairstyles, \additionalNumber additional attributes, and hair segmentation masks. We show the effectiveness of our dataset in various hairstyle-related tasks, such as hair dyeing, hairstyle transfer, and hairstyle classification.
\section{K-Hairstyle}

\begin{figure}[t!]
    \centering
    \includegraphics[width=\linewidth]{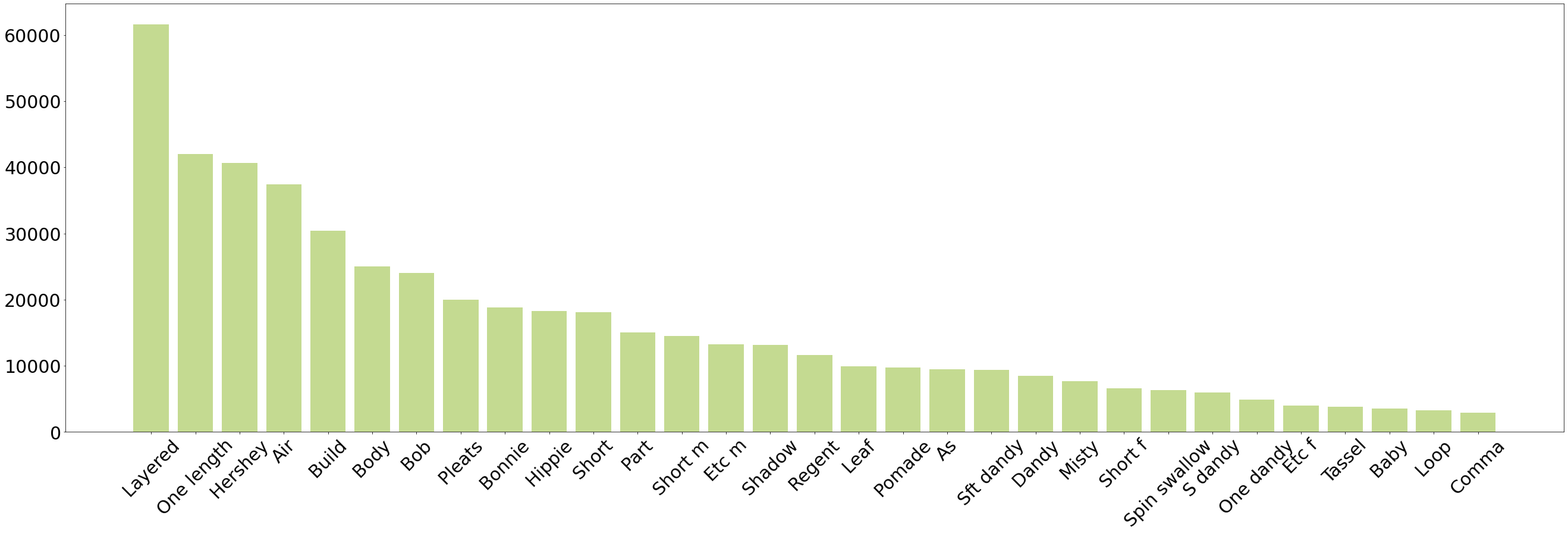}
    \vspace{-0.7cm}
    \caption{The number of images per hairstyle class.}
    \vspace{-0.5cm}
    \label{dist:basestyle}
\end{figure}

\subsection{Data description}

\dataset provides \datasetNumber high-resolution images with a rich set of annotations, such as hairstyle classes, hair segmentation masks, and various attributes, the details of which are as follows: 
\begin{itemize}[leftmargin=4.5mm, noitemsep, topsep=0pt]
    \item \textbf{High-resolution image.} The images with a resolution of 4032$\times$3024 are collected using high-end cameras.
    \item \textbf{Large-scale dataset.} We provide \datasetNumber images, more than any other existing hairstyle datasets.
    \item \textbf{Multi-view image.} The dataset contains multi-view images that are captured from various camera angles for each person. The angles include two different vertical camera angles and about ten to sixty different horizontal angles.
    \item \textbf{Hair segmentation mask.} The hair regions of images are manually labeled in the form of a polygon.
    \item \textbf{Hairstyle attributes.} Various hairstyle-related attributes are annotated by Korean expert hairstylists. In detail, different hairstyles are categorized into \classNumber types, and \additionalNumber additional attributes, such as hair color, length, and curl, are also labeled.
    \item \textbf{Blurred face.} Due to the privacy issue, we made the facial region blurry.
\end{itemize}

\begin{figure*}[ht!]
    \centering
    \includegraphics[width=\linewidth]{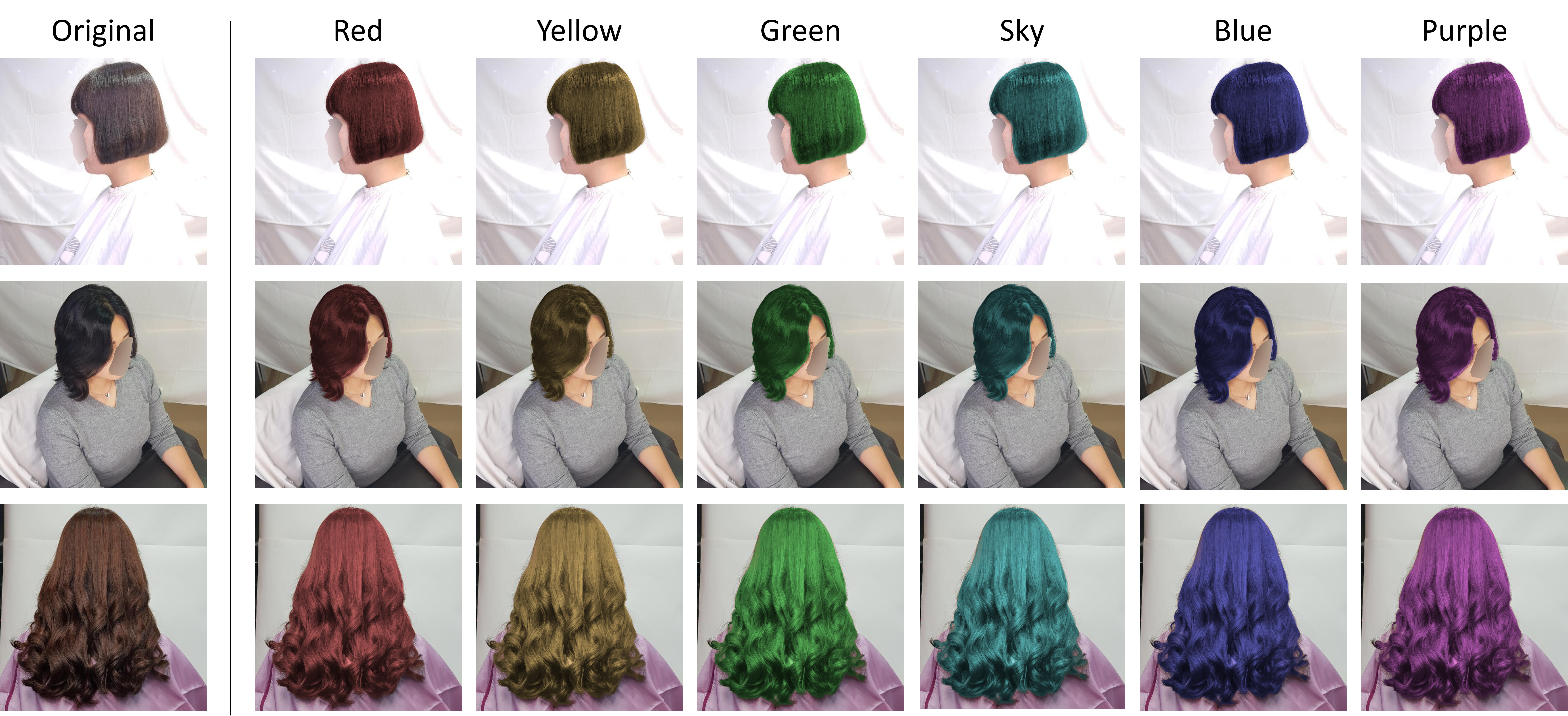}
    \vspace{-0.5cm}
    \caption{Hair dyeing results.}
    \vspace{-0.4cm}
    \label{fig:hair_dyeing}
\end{figure*}

\vspace{-0.2cm}
\subsection{Data collection}
In this section, we describe how we collected the image data and its annotation labels including hairstyle attributes and hair segmentation masks.

\noindent\textbf{Image Collection.}
We first set the standard categories of Korean hairstyles with beauty and arts professors, the representatives of Korean beauty brands, and well-known hairstylists. 
We define \classNumber types of hairstyles which include common Korean hairstyles.
Then, we collect 400,000 images (a fourth of total images) containing those hairstyles through 300 expert hairstylists.
The stylists are asked to take photos of each hairstyle from two different vertical camera angles (high and middle) and at least 10 different horizontal camera angles, changing the angle by 6 to 36 degrees each time. 

Moreover, we collect 100,000 images (a fifth of total images) via crowd-sourcing platform. After obtaining sufficient number of images using crowd-sourcing, we manually filtered the images which satisfy the conditions mentioned above (e.g., whether the images are in multi-view or the images contain appropriate hairstyles). There is no significant difference between the images collected by hairstylists and crowd-sourcing.

\noindent\textbf{Segmentation Masks and Attribute Annotation.}
We annotate hair segmentation masks with the help of an auto-labeling model and human labelers. The auto-labeling model first annotates each mask, and human labelers manually refine the masks. In addition, the expert hairstyle experts label various hairstyle attributes, such as the type of hairstyle, hair length, curl strength, etc. Figure~\ref{dist:basestyle} shows the distribution of the \classNumber hairstyle classes in our dataset.


\section{K-Hairstyle applications}
\vspace{-0.2cm}
In this section, we demonstrate the effectiveness of our dataset via three applications: hair dyeing, hairstyle transfer, and hairstyle classification. For these experiments, we randomly sampled 130,000 images from \dataset. Then, we divided the images into \trainNumber training, \validationNumber validation, and \testNumber test images. 

\vspace{-0.2cm}
\subsection{Hair Dyeing}

To dye the hair, it is necessary to predict segmentation masks corresponding to hair regions. With \dataset, we train and evaluate the five segmentation models, including U-Net~\cite{ronneberger2015u}, Feature Pyramid Network (FPN)~\cite{lin2017feature}, Pyramid Scene Parsing Network~\cite{zhao2017pyramid}, DeepLabv3+~\cite{chen2018encoder}, and Pyramid Attention Network (PAN)~\cite{li2018pyramid}.
We evaluate the performance of the segmentation models using three widely-used image segmentation measures, including mean Intersection over Union (mIoU), F1 score, and accuracy.
Table~\ref{Table:segmentation} shows these results, the number of Multiply–ACcumulate operations (MACs) measured with respect to a
512$\times$512 input image, and the required memory size to accommodate the parameters of each model (\#Params).
PAN~\cite{li2018pyramid} shows the best performances compared to other baselines, and PSPNet~\cite{zhao2017pyramid} is shown to be the most efficient in terms of the number of parameters and computational operations.

\begin{figure}[t!]
    \centering
    \includegraphics[width=\linewidth]{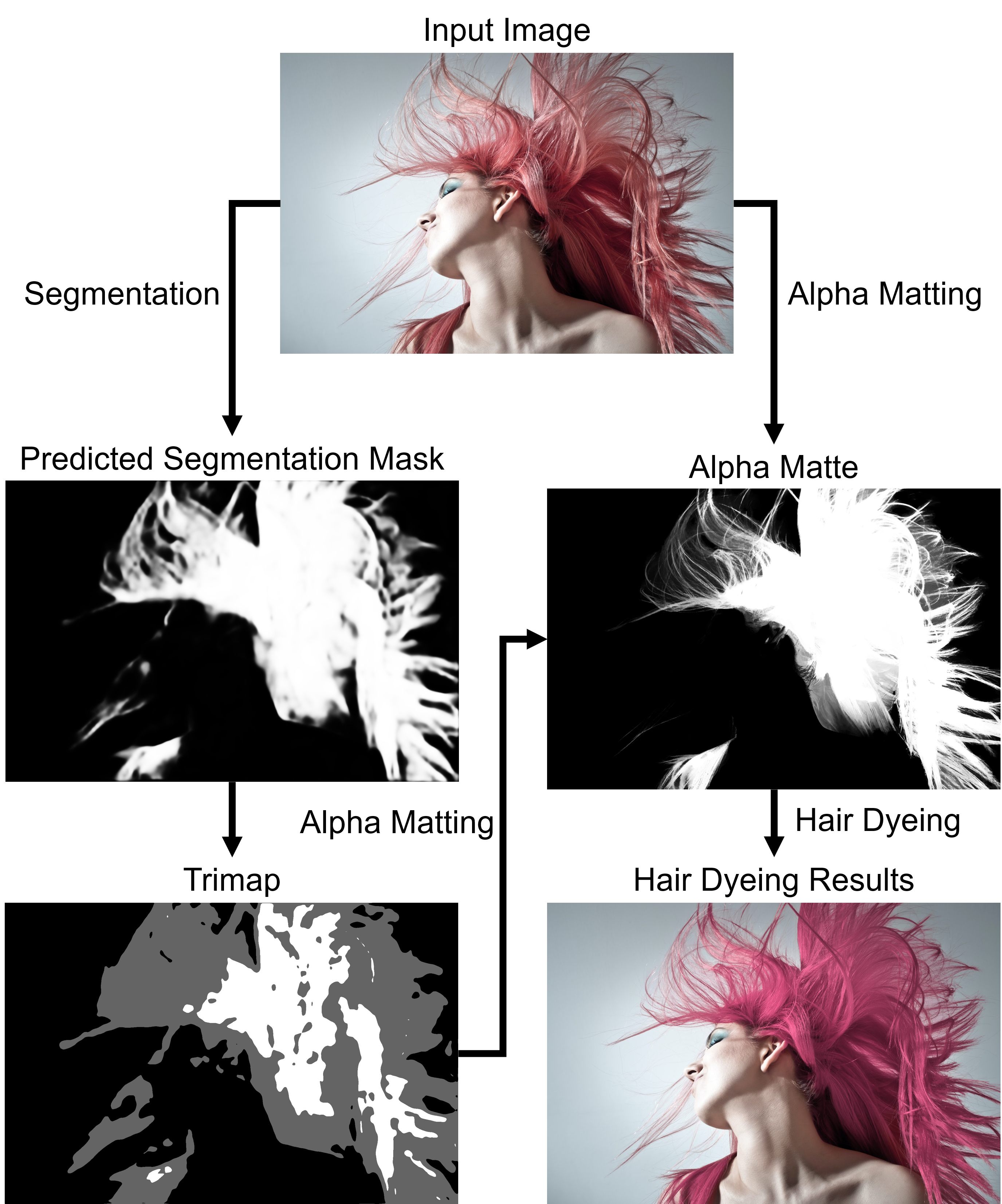}
    \vspace{-0.5cm}
    \caption{Overview of our hair dyeing approach.}
    \vspace{-0.8cm}
    \label{model:hair_coloring}
\end{figure}

\begin{figure*}[t!]
    \centering
    \includegraphics[width=\linewidth]{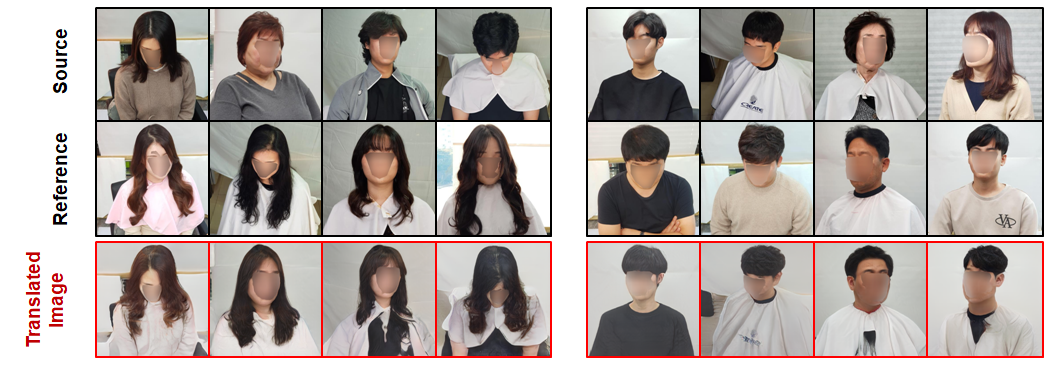}
    \vspace{-0.8cm}
    \caption{Hair transfer results.}
    \vspace{-0.5cm}
    \label{fig:translation}
\end{figure*} 

\begin{table}[t!]
    \centering
    \small
    \begin{tabular}{@{}p{0.09\textwidth}<{\centering}|p{0.05\textwidth}<{\centering}p{0.05\textwidth}<{\centering}p{0.07\textwidth}<{\centering}|p{0.04\textwidth}<{\centering}p{0.065\textwidth}<{\centering}@{}}
        \toprule
        Model & mIoU$_{\uparrow}$ & F1$_{\uparrow}$ & Accuracy$_{\uparrow}$ & MACs$_{\downarrow}$ & \#Params$_{\downarrow}$\\
        \midrule
        U-Net & 0.9182 & 0.9569 & 0.9862 & 62.12 & 51.5\\
        FPN & 0.9182 & 0.9570 & 0.9862 & 50.71 & 45.1 \\
        PSPNet & 0.9171 & 0.9564 & 0.9859 & \textbf{11.74} & \textbf{2.2} \\
        DeepLabv3+ & 0.9185 & 0.9571 & 0.9862 & 56.20 & 45.6\\
        PAN & \textbf{0.9187} & \textbf{0.9573} & \textbf{0.9863} & 54.22 & 43.2 \\
        \bottomrule
    \end{tabular}
    \vspace{-0.4cm}
    \caption{Comparison of segmentation performances.}
    \vspace{-0.5cm}
    \label{Table:segmentation}
\end{table}

As shown in Figure~\ref{model:hair_coloring}, after predicting the hair segmentation mask of the input image, we conduct alpha matting and hair dyeing. We leverage k-nearest-neighbor matting~\cite{chen2013knn}, an alpha matting approach that provides accurate segmentation masks.
Alpha matting accurately divides an input image into the foreground (hair) and background (the rest) with a trimap which divides the entire region into three parts, based on the predicted segmentation mask.
The foreground, the background, and the unknown part in the trimap are visualized in black, white, and gray, respectively, in Figure~\ref{model:hair_coloring}.
With the accurate segmentation masks obtained in this manner, we generate the photo-realistic hair dyeing images by modifying HSV instead of RGB. Figure~\ref{fig:hair_dyeing} describes hair dyeing results.

\vspace{-0.2cm}
\subsection{Hairstyle Transfer}
We also perform hairstyle transfer as another application of \dataset. 
We aim to transfer the hairstyle of an input image to the desired hairstyle in a reference image, regardless of their gender or viewing angles. Therefore, we employ a reference-based image-to-image transfer model, StarGAN v2~\cite{choi2020starganv2}.
StarGAN v2 transfers the style of an input image to the target style extracted from a reference image, while maintaining other features (e.g., pose) of the input image. When extracting the style from a reference image, StarGAN v2 utilizes the domain-specific style encoders to reflect the distinctive features of the corresponding domain. For the experiment, we split the images into two domains, male and female, to reflect the difference in each gender's common hairstyles. Also, we train and test the model using 512$\times$512 cropped images.

As presented in Figure~\ref{fig:translation}, we successfully transfer the hairstyles of source images to those of reference images, even when the gender or the viewing angles are different. To evaluate the quality of the translated images, we measure the FID score~\cite{heusel2017fid} between the real images from the training data and the translated images. We achieved 14.67 for the transfer to the male domain, and 14.68 for the female domain, with the average value of 14.67.


\subsection{Hairstyle Classification}

\begin{table}[t!]
    \centering
    \small
    \begin{tabular}{ccccc}
        \toprule
        Model & Accuracy$_{\uparrow}$ & \#Attribute & Augmentation & Crop \\
        \midrule
        ResNet18 & 0.3119 & 27 & X & X \\ 
        ResNet18 & 0.3189 & 27 & O & X \\ 
        ResNet18 & 0.3333 & 27 & O & O \\
        ResNet18 & \textbf{0.3792} & \textbf{51} & O & O \\ 
        \bottomrule
    \end{tabular}
    \vspace{-0.2cm}
    \caption{Ablation study of classification models.}
    \vspace{-0.2cm}
    \label{Table:classification2}
\end{table}

\begin{table}[h!]
    \centering
    \small
    \begin{tabular}{cccc}
        \toprule
        Model & Accuracy$_{\uparrow}$ & MACs(B)$_{\downarrow}$ & \#Params(M)$_{\downarrow}$\\
        \midrule
        ResNet18 & 0.3792 & \textbf{9.50} & \textbf{11.20} \\
        ResNet50 & 0.3699 & 21.47 & 23.61 \\
        ResNet101 & \textbf{0.3880} & 40.91 & 42.60 \\
        Wide-ResNet50 & 0.3764 & 59.69 & 66.93 \\ 
        \bottomrule
    \end{tabular}
    \vspace{-0.2cm}
    \caption{Comparison of classification performances.}
    \vspace{-0.5cm}
    \label{Table:classification}
\end{table}


We conduct hairstyle classification with \dataset. As shown in Table~\ref{Table:classification2}, we perform an ablation study to analyze the effects of the number of attributes, data augmentation while training, and 224$\times$224 cropped images. The data augmentation contains common techniques such as horizontal flip, rotation, and distortion.
The accuracy increases significantly when training with the additional attributes, such as length, gender, and angle.
Also, among ResNet18, ResNet50, ResNet101~\cite{he2016deep}, and Wide-ResNet50~\cite{zagoruyko2016wide}, ResNet101~\cite{he2016deep} achieves the highest classification accuracy on \dataset. 


\vspace{-0.2cm}
\section{Conclusion}

In this paper, we introduce \dataset dataset, a novel large-scale Korean hairstyle dataset containing segmentation masks and various attribute labels.
To demonstrate the effectiveness of our dataset, we present three applications, hair dyeing, hairstyle transfer, and hairstyle classification.
We believe \dataset dataset will open up a wide range of research opportunities in the area of hairstyle editing and manipulation.

\textbf{Acknowledgments}. 
This work was partially supported by National Information Society Agency (NIA) (No.2020-0561-54) and by Institute of Information \& communications Technology Planning \& Evaluation (IITP) grant funded by the Korea government (MSIT) (No.2019-0-00075, Artificial Intelligence Graduate School Program (KAIST)).




\clearpage

\bibliographystyle{IEEEbib}
\bibliography{strings,refs}

\end{document}